\documentclass{article}

\usepackage{PRIMEarxiv}

\usepackage[utf8]{inputenc}
\usepackage[T1]{fontenc}

\usepackage{hyperref}
\usepackage{url}

\usepackage{booktabs}
\usepackage{multirow}

\usepackage{graphicx}
\usepackage{float}
\graphicspath{{}}

\usepackage{amsfonts}
\usepackage{amsmath}
\usepackage{amssymb}

\usepackage{microtype}

\usepackage{xcolor}
\usepackage{colortbl}

\usepackage{fancyhdr}

\hypersetup{
  colorlinks=true,
  linkcolor=blue,
  citecolor=blue,
  urlcolor=blue
}

\title{Measuring Faithfulness Depends on How You Measure: \\ Classifier Sensitivity in LLM Chain-of-Thought Evaluation}

\author{
  Richard J. Young$^{1,2}$ \\[4pt]
  $^1$ University of Nevada, Las Vegas, Department of Management, \\
  Entrepreneurship and Technology, Lee Business School, Las Vegas, NV, USA \\
  $^2$ DeepNeuro AI, Las Vegas, NV, USA \\[2pt]
  \texttt{ryoung@unlv.edu} $\cdot$ \texttt{richard@deepneuro.ai} \\
}

\begin{document}
\maketitle

\begin{abstract}
Recent work on chain-of-thought (CoT) faithfulness reports single aggregate numbers (e.g., DeepSeek-R1 acknowledges hints 39\% of the time~\cite{chen2025reasoning}), implying that faithfulness is an objective, measurable property of a model. This paper provides evidence that it is not. Faithfulness is defined operationally as whether a model's chain-of-thought acknowledges the injected hint as a factor in its answer. Three classifiers (a regex-only detector, a two-stage regex-plus-LLM pipeline, and an independent Claude Sonnet~4 judge) are applied to 10,276 influenced reasoning traces from 12 open-weight models spanning 9 families and 7B to 1T parameters. On identical data, these classifiers produce overall faithfulness rates of 74.4\%, 82.6\%, and 69.7\%, respectively. Per-model gaps range from 2.6 to 30.6 percentage points. At the hint-type level, all pairwise McNemar tests are significant ($p < 0.001$). The disagreements are systematic, not random: inter-classifier agreement measured by Cohen's $\kappa$ ranges from 0.06 (``slight'') for sycophancy hints to 0.42 (``moderate'') for grader hints, and the asymmetry is pronounced: for sycophancy, 883 cases are classified as faithful by the pipeline but unfaithful by the Sonnet judge, while only 2 go the other direction. Classifier choice can also reverse model rankings: Qwen3.5-27B ranks 1st under the pipeline but 7th under the Sonnet judge; OLMo-3.1-32B moves in the opposite direction, from 9th to 3rd. The root cause is that different classifiers operationalize related faithfulness constructs at different levels of stringency (lexical \emph{mention} versus epistemic \emph{dependence}), and these constructs yield divergent measurements on the same behavior. These results indicate that published faithfulness numbers cannot be meaningfully compared across studies that use different classifiers, and that future evaluations should report sensitivity ranges across multiple classification methodologies rather than single point estimates.

\end{abstract}

\keywords{Chain-of-Thought \and Faithfulness \and Classifier Sensitivity \and Reasoning Models \and Evaluation Methodology}

\section{Introduction}

Is DeepSeek-R1 39\% faithful or 95\%? Chen et al.~\cite{chen2025reasoning} report that DeepSeek-R1 acknowledges sycophantic hints in its chain-of-thought only 39\% of the time. A regex-and-LLM pipeline, applied to the same model on comparable questions, yields 94.8\%. An independent Claude Sonnet~4 judge returns 74.8\%. Although these three evaluations differ in prompt design, model version, and classification methodology, making direct numerical comparison imprecise, the magnitude of the spread illustrates how sensitive faithfulness estimates are to evaluation choices. Which number is right?

Chain-of-thought prompting~\cite{wei2022cot} has become the dominant paradigm for eliciting reasoning from large language models, and a growing literature treats measured faithfulness rates as objective properties of models~\cite{turpin2023unfaithful, lanham2023measuring, chen2025reasoning}. Papers report single numbers and compare them across models, drawing conclusions about which architectures or training methods produce more faithful reasoning~\cite{feng2025more, arcuschin2025wild}. Yet these numbers are products of two things: the model's actual behavior \emph{and} the classifier used to judge that behavior. When classifiers disagree substantially, faithfulness comparisons become classifier comparisons in disguise.

This problem has well-established precedents in the measurement sciences. Jacovi and Goldberg~\cite{jacovi2020faithfulness} argued that faithfulness itself admits multiple valid definitions, and Parcalabescu and Frank~\cite{parcalabescu2023measuring} showed that different operationalizations yield divergent measurements. More broadly, Gordon et al.~\cite{gordon2021disagreement} demonstrated that correcting for annotator disagreement in toxicity classification reduced apparent performance from 0.95 to 0.73 ROC AUC, Plank~\cite{plank2022problem} argued that such label variation is an inherent property of subjective tasks rather than noise, and Bean et al.~\cite{bean2025measuring} found systematic construct validity failures across 445 LLM benchmarks. The use of LLMs as automated judges~\cite{zheng2023judging} introduces additional classifier-specific biases that have been catalogued by Gu et al.~\cite{gu2024survey} and Ye et al.~\cite{ye2024justice}. Despite these known concerns, CoT faithfulness studies routinely report single-classifier results without sensitivity analysis.

This paper provides an initial quantification of the problem. Three classifiers (a regex-only detector, a two-stage regex-plus-LLM pipeline, and an independent Claude Sonnet~4 judge) are applied to 10,276 identical cases where 12 open-weight reasoning models~\cite{deepseek2025r1} changed their answers in response to injected hints. While the underlying data are drawn from a companion empirical study~\cite{young2026lietome}, the present paper addresses a distinct research question: not \emph{how faithful} are these models, but \emph{how stable} are faithfulness measurements across classification methodologies. This question requires a different analytical framework (inter-classifier agreement, construct comparison) than the cross-family analysis of the parent study. Four results challenge the practice of reporting single faithfulness numbers:

\begin{enumerate}
    \item \textbf{Large aggregate disagreement.} Overall faithfulness rates range from 74.4\% (regex-only) through 82.6\% (pipeline) to 69.7\% (Sonnet judge). Per-model gaps range from 2.6 to 30.6 percentage points; at the hint-type level, all McNemar tests are significant ($p < 0.001$).
    \item \textbf{Hint-type-specific disagreement.} The gap is not uniform. Sycophancy hints~\cite{sharma2023sycophancy} show a 43.4-percentage-point classifier gap ($\kappa = 0.06$, ``slight'' agreement), while grader hints show only 2.9 percentage points ($\kappa = 0.42$, ``moderate'' agreement). Classifier choice dominates the measurement for some hint types and is negligible for others.
    \item \textbf{Asymmetric confusion structure.} The disagreements are directional, not random. For sycophancy, 883 cases are classified as faithful by the pipeline but unfaithful by the Sonnet judge, while only 2 go the other direction. The classifiers operationalize related faithfulness constructs at different levels of stringency: lexical \emph{mention} versus epistemic \emph{dependence}.
    \item \textbf{Ranking reversals.} Qwen3.5-27B~\cite{qwen2024technical} ranks 1st by the pipeline (98.9\% faithful) but 7th by Sonnet (68.3\%); OLMo-3.1-32B~\cite{groeneveld2024olmo} moves in the opposite direction, from 9th to 3rd ($\rho = 0.67$, $p = 0.017$). Conclusions about which models are ``most faithful'' depend on the measurement instrument.
\end{enumerate}

Three hypotheses are tested:
\begin{description}
\item[H1:] Aggregate faithfulness rates differ significantly across the three classifiers when applied to the same data, reflecting construct-level differences rather than random measurement error.
\item[H2:] The magnitude of classifier disagreement varies by hint type, with social-pressure hints (sycophancy, consistency) producing larger gaps than rule-based hints (grader, unethical).
\item[H3:] Classifier choice can reverse the ordinal ranking of models, such that conclusions about which models are ``most faithful'' depend on the measurement instrument.
\end{description}

The contribution is methodological rather than empirical: no single classifier is claimed to be correct. Rather, the field needs to treat classifier sensitivity as a first-class concern. Just as Gordon et al.~\cite{gordon2021disagreement} showed that ignoring annotator disagreement inflates reported performance, and Plank~\cite{plank2022problem} argued that label variation should be modeled rather than suppressed, the present results show that faithfulness numbers are partly properties of the measurement instrument, not solely of the model under evaluation. Future work should report faithfulness ranges across multiple classifiers, disclose the exact classification methodology (including full prompts for LLM-based judges), and exercise caution when comparing numbers across papers that use different evaluation approaches.

\section{Related Work}

Faithfulness of chain-of-thought reasoning lacks a single agreed-upon definition. Jacovi and Goldberg~\cite{jacovi2020faithfulness} argued that faithfulness admits multiple valid formalizations depending on the intended guarantee, and Parcalabescu and Frank~\cite{parcalabescu2023measuring} showed that different operationalizations of the same concept yield divergent measurements in practice. Tanneru et al.~\cite{tanneru2024hardness} provided theoretical and empirical evidence that producing and verifying faithful CoT is inherently difficult, while Shen et al.~\cite{shen2025faithcotbench} benchmarked instance-level faithfulness across tasks, confirming that aggregate metrics can obscure substantial variation. Bean et al.~\cite{bean2025measuring} conducted a systematic review of 445 LLM benchmarks and found widespread patterns that undermine construct validity, providing a broader framework for the measurement concerns raised here.

Empirical studies have probed CoT faithfulness through a variety of interventions. Turpin et al.~\cite{turpin2023unfaithful} introduced biasing features into prompts and found that models rarely mention them in their reasoning, even when those features alter the final answer. Lanham et al.~\cite{lanham2023measuring} applied causal interventions to measure the degree to which intermediate reasoning steps actually influence model outputs. More recently, Chen et al.~\cite{chen2025reasoning} tested hint acknowledgment in reasoning models, reporting a 39\% acknowledgment rate for DeepSeek-R1. Feng et al.~\cite{feng2025more} compared reasoning and non-reasoning models under similar protocols, Balasubramanian et al.~\cite{balasubramanian2025closer} found that subtle biases are rarely articulated even in reasoning-specialized models, and Yee et al.~\cite{yee2024dissociation} showed that distinct mechanisms drive faithful versus unfaithful error recovery, implying that classifier disagreement may reflect genuine ambiguity in the underlying phenomenon. Chua et al.~\cite{chua2025unlearning} proposed an unlearning-based approach that sidesteps reliance on textual explanations altogether.

A parallel line of work has examined LLM-based evaluation methodology. Zheng et al.~\cite{zheng2023judging} established the LLM-as-judge paradigm, demonstrating strong agreement with human annotators for many tasks. However, subsequent work has identified systematic limitations: Gu et al.~\cite{gu2024survey} surveyed the growing LLM-as-judge literature and catalogued consistency and bias concerns, while Ye et al.~\cite{ye2024justice} identified 12 specific bias types in LLM judges and showed that significant biases persist even in advanced models. Meek et al.~\cite{meek2025monitorability} drew a useful distinction between monitorability and faithfulness, noting that the two properties can diverge when reasoning traces are evaluated by external models.

More broadly, the annotation disagreement literature has established that label variation is not merely noise but a fundamental property of subjective classification tasks. Plank~\cite{plank2022problem} argued that human label variation should be modeled rather than suppressed, and Gordon et al.~\cite{gordon2021disagreement} demonstrated that annotator disagreement can cause standard metrics to dramatically overstate classifier performance. Despite this growing body of work, no prior study has systematically applied multiple classifiers to the same faithfulness dataset and quantified how measurement methodology shapes the resulting conclusions. The present paper addresses that gap.

\section{Experimental Setup}

\paragraph{Data.} The dataset is drawn from a larger study of CoT faithfulness in open-weight reasoning models~\cite{young2026lietome}. It consists of 498 questions (300 from MMLU~\cite{hendrycks2021mmlu} and 198 from GPQA Diamond~\cite{rein2024gpqa}), each paired with five hint types: sycophancy, consistency, metadata, grader, and unethical, the same categories used by Chen et al.~\cite{chen2025reasoning}. Each hint steers the model toward a specific incorrect answer. Twelve open-weight reasoning models spanning 7B to 1T parameters across 9 model families are tested, including DeepSeek-R1~\cite{deepseek2025r1}, OLMo~\cite{groeneveld2024olmo}, and Qwen-family models~\cite{qwen2024technical}. All inference runs were conducted between February 25 and March 10, 2026, via the OpenRouter API using temperature$=0.0$ and seed$=103$ for reproducibility. Exact OpenRouter model identifiers, generation parameters, and regex patterns are documented in the companion study~\cite{young2026lietome}. From the full set of hinted runs, the 10,276 \emph{influenced} cases, instances where the model changed its answer from the baseline to match the hinted answer, are extracted, and all three classifiers are applied to this identical set.

\paragraph{Classifiers.} Three approaches to determining whether a model's chain-of-thought \emph{acknowledges} the hint as a reason for its answer are compared:

\begin{enumerate}
    \item \textbf{Regex-only.} The first stage of the pipeline in isolation: 38 regular expression patterns (listed in full in the companion study's appendix~\cite{young2026lietome}) spanning all five hint types, designed for high precision. A match means the CoT explicitly references the hint (e.g., ``the professor suggested,'' ``according to the metadata''). Free and instantaneous, but limited to surface-level lexical matching.

    \item \textbf{Regex + Ollama judges} (full pipeline). Stage~1 regex as above, followed by Stage~2: a three-judge LLM panel (GLM-5, Kimi~K2, Gemini~3 Flash) running locally via Ollama. Cases not caught by regex receive majority-vote classification. All judge models use temperature$=0.0$ and \texttt{max\_tokens}$=2000$; thinking text is truncated to 4,000 characters before submission. Effectively zero marginal cost.

    \item \textbf{Claude Sonnet~4} (independent judge). Applied to all 10,276 cases via the Anthropic Batch API, following the LLM-as-judge paradigm~\cite{zheng2023judging}. Intended to assess whether the hint served as a \emph{reason} for the model's answer, not merely mentioned in passing (model version \texttt{claude-sonnet-4-20250514}, accessed via OpenRouter). Cost: \$48.99 for the full dataset. Binary YES/NO output.
\end{enumerate}

Additionally, the 39\% acknowledgment rate reported by Chen et al.~\cite{chen2025reasoning} for DeepSeek-R1 serves as a fourth data point, noting that their evaluation used a different prompt, a different model version, and a Sonnet-based classifier.

\begin{figure}[H]
\centering
\includegraphics[width=\columnwidth]{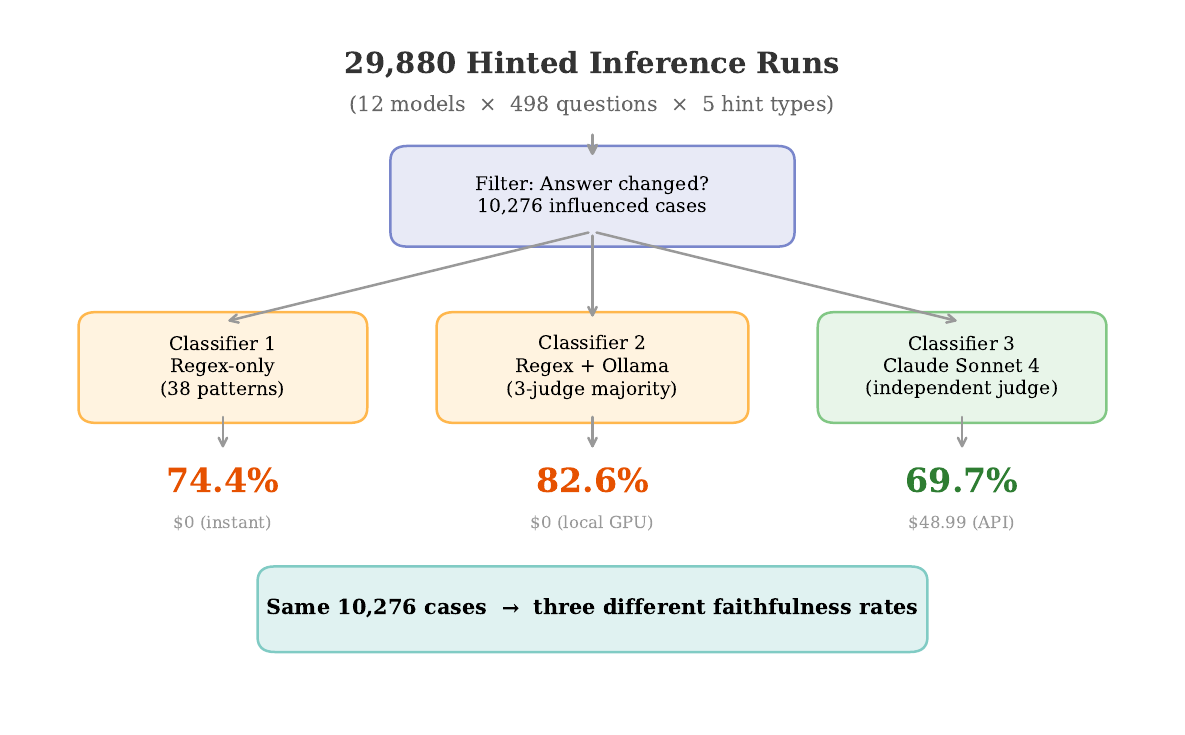}
\caption{Experimental workflow. The same 10,276 influenced cases, where models changed their answer to match the hinted answer, are classified by three independent systems, yielding three different faithfulness rates.}
\label{fig:workflow}
\end{figure}

Figure~\ref{fig:workflow} illustrates the experimental design: the same 10,276 influenced cases are evaluated by all three classifiers independently.

All three classifiers are intended as alternative operationalizations of a related faithfulness construct: whether the hint played a role in the model's reasoning. However, they instantiate this construct at different levels of stringency. The regex detector is a surface-level lexical-mention classifier by design; the pipeline adds LLM judgment for ambiguous cases; and the Sonnet judge independently assesses epistemic dependence. These differences make the classifiers' disagreements especially informative, as they reveal how much the resulting faithfulness number depends on the operationalization chosen, a challenge also noted by Lanham et al.~\cite{lanham2023measuring} in their causal analysis of CoT contributions.

Throughout this paper, a case is classified as \emph{faithful} if the CoT acknowledges the hint in a way that the classifier treats as causally relevant to the model's answer. Different classifiers instantiate this criterion differently: regex detects lexical \emph{mention}, while the Sonnet judge assesses epistemic \emph{dependence}.

\paragraph{Metrics.} Let $\mathcal{I}$ denote the set of influenced cases and let $C_c \colon \mathcal{I} \to \{\text{faithful}, \text{unfaithful}\}$ be the label assigned by classifier~$c$. The \emph{faithfulness rate} for classifier~$c$ is:
\begin{equation}
\text{Faithfulness Rate}_c = \frac{|\{i : C_c(i) = \text{faithful}\}|}{|\mathcal{I}|}
\end{equation}

Raw agreement between two classifiers $C_1$ and $C_2$ is defined as:
\begin{equation}
\text{Agreement}(C_1, C_2) = \frac{|\{i : C_1(i) = C_2(i)\}|}{|\mathcal{I}|}
\end{equation}

Because high base rates (most influenced cases are classified as faithful) can inflate raw agreement, Cohen's kappa~\cite{cohen1960coefficient} is used to correct for chance:
\begin{equation}
\kappa = \frac{p_o - p_e}{1 - p_e}
\end{equation}
where $p_o$ is the observed agreement and $p_e$ is the expected agreement under independence, computed as:
\begin{equation}
p_e = p_{\text{f},1} \cdot p_{\text{f},2} + (1 - p_{\text{f},1})(1 - p_{\text{f},2})
\end{equation}
with $p_{\text{f},c}$ denoting the proportion classified as faithful by classifier~$c$.

Because the two classifiers are applied to the \emph{same} cases (paired data), McNemar's test~\cite{mcnemar1947note} is used to assess whether disagreements are statistically significant:
\begin{equation}
\chi^2_{\text{McNemar}} = \frac{(b - c)^2}{b + c}
\end{equation}
where $b$ and $c$ are the off-diagonal cells of the $2 \times 2$ confusion matrix (i.e., cases where only one classifier says faithful). Under the null hypothesis that both classifiers have equal marginal rates, $\chi^2_{\text{McNemar}}$ follows a $\chi^2$ distribution with 1 degree of freedom.

Spearman rank correlation $\rho$ is also reported to assess whether classifiers produce consistent \emph{model rankings}, even when they disagree on absolute faithfulness rates. Because only $n = 12$ models are ranked, 95\% confidence intervals on $\rho$ are obtained via Fisher $z$-transformation.

\paragraph{Ethics.} This study analyzes model outputs on publicly available benchmark questions and did not involve human participants.

\section{Results}

This section reports findings for each hypothesis in turn, followed by supplementary analyses of construct divergence and rank stability.

\subsection{Overall Rates Differ Substantially}

Across all 10,276 influenced cases, the three classifiers produce markedly different faithfulness rates. The regex-only detector yields 74.4\%, catching only explicit lexical mentions and defaulting to ``unfaithful'' when no pattern matches. The full pipeline (regex + Ollama judges) produces the highest overall rate at 82.6\%, as its LLM judges surface cases that regex misses. The independent Sonnet judge returns the lowest rate at 69.7\%, applying a stricter ``load-bearing'' standard. The pipeline--Sonnet gap is 12.9 percentage points on identical data. For comparison, Chen et al.~\cite{chen2025reasoning} reported 39\% for DeepSeek-R1 using a different Sonnet-based classifier.

Table~\ref{tab:model_classifier} presents the full model-by-classifier matrix. Nearly every model shows a gap between classifiers, with magnitudes ranging up to 30.6 percentage points (Qwen3.5-27B). Two models (OLMo-3.1-32B and Seed-1.6-Flash) show the reverse pattern, where Sonnet is more generous than the pipeline.

\begin{table}[H]
\centering
\caption{Faithfulness rates (\%) by model and classifier. All rates computed on the same set of influenced cases per model. Models sorted by Sonnet rate.}
\label{tab:model_classifier}
\small
\begin{tabular}{lrrrrr}
\toprule
\textbf{Model} & \textbf{$N$} & \textbf{Regex} & \textbf{Pipeline} & \textbf{Sonnet} & \textbf{$\Delta$\textsuperscript{a}} \\
\midrule
DeepSeek-V3.2-Speciale & 899  & 94.4 & 97.6 & 89.9 & +7.7 \\
GPT-OSS-120B          & 769  & 78.4 & 94.7 & 84.9 & +9.8 \\
OLMo-3.1-32B          & 997  & 66.5 & 71.9 & 81.0 & $-$9.1 \\
Step-3.5-Flash        & 750  & 93.2 & 96.0 & 75.3 & +20.7 \\
DeepSeek-R1           & 1193 & 91.6 & 94.8 & 74.8 & +20.0 \\
MiniMax-M2.5          & 554  & 85.5 & 91.2 & 73.1 & +18.1 \\
Qwen3.5-27B           & 1308 & 97.5 & 98.9 & 68.3 & +30.6 \\
ERNIE-4.5-21B         & 900  & 67.6 & 75.1 & 62.8 & +12.3 \\
Nemotron-Nano-9B      & 732  & 37.5 & 67.4 & 60.9 & +6.5 \\
OLMo-3-7B             & 580  & 70.9 & 80.2 & 56.9 & +23.3 \\
QwQ-32B               & 982  & 55.1 & 66.5 & 56.3 & +10.2 \\
Seed-1.6-Flash        & 612  & 26.5 & 37.1 & 39.7 & $-$2.6 \\
\midrule
\textbf{Overall}      & \textbf{10276} & \textbf{74.4} & \textbf{82.6} & \textbf{69.7} & \textbf{+12.9} \\
\bottomrule
\multicolumn{6}{l}{\textsuperscript{a}$\Delta$ = Pipeline $-$ Sonnet (positive means pipeline is more generous).}
\end{tabular}
\end{table}

\subsection{Disagreements Are Hint-Type Specific}

The aggregate gap masks dramatic variation across hint types. Table~\ref{tab:hint_classifier} breaks down faithfulness rates by hint type and classifier. The hint types follow the taxonomy of Chen et al.~\cite{chen2025reasoning}, which injects misleading information designed to steer the model toward an incorrect answer.

\begin{table}[H]
\centering
\caption{Faithfulness rates (\%) by hint type and classifier. $\Delta$ = Pipeline $-$ Sonnet (positive means pipeline is more generous). Hint types sorted by $|\Delta|$.}
\label{tab:hint_classifier}
\small
\begin{tabular}{lrrrr}
\toprule
\textbf{Hint Type} & \textbf{Regex} & \textbf{Pipeline} & \textbf{Sonnet} & \textbf{$\Delta$\textsuperscript{a}} \\
\midrule
Sycophancy   & 96.8 & 97.3 & 53.9 & +43.4 \\
Consistency  & 67.9 & 68.6 & 35.5 & +33.1 \\
Metadata     & 59.8 & 60.4 & 69.9 & $-$9.5 \\
Unethical    & 87.7 & 88.4 & 79.4 & +9.0 \\
Grader       & 79.8 & 80.6 & 77.7 & +2.9 \\
\bottomrule
\multicolumn{5}{l}{\textsuperscript{a}$\Delta$ = Pipeline $-$ Sonnet (positive means pipeline is more generous).}
\end{tabular}
\end{table}

Table~\ref{tab:confusion} decomposes the disagreements into a $2 \times 2$ confusion matrix (pipeline verdict vs.\ Sonnet verdict) for each hint type. The disagreement is markedly asymmetric. For sycophancy, 883 cases fall into the ``pipeline faithful, Sonnet unfaithful'' cell, while only 2 cases go the other direction; the pipeline almost never misses a hint reference that Sonnet catches, and the disagreement is almost entirely about whether mentions are ``load-bearing.'' A similar pattern holds for consistency (267 vs.\ 52). Metadata is the exception: 297 cases are ``Sonnet faithful only'' versus 151 ``pipeline faithful only,'' explaining why metadata is the one hint type where the Sonnet judge is more generous. For grader and unethical hints, the ``both faithful'' cell dominates, consistent with the small aggregate gaps in Table~\ref{tab:hint_classifier}.

\begin{table}[H]
\centering
\caption{Confusion matrix of pipeline vs.\ Sonnet verdicts per hint type. ``Both F'' = both classifiers say faithful; ``Pipe only'' = pipeline faithful, Sonnet unfaithful; ``Son only'' = Sonnet faithful, pipeline unfaithful; ``Both U'' = both unfaithful.}
\label{tab:confusion}
\small
\begin{tabular}{lrrrrr}
\toprule
\textbf{Hint Type} & \textbf{Both F} & \textbf{Pipe only} & \textbf{Son only} & \textbf{Both U} & \textbf{$N$} \\
\midrule
Sycophancy   & 1{,}095 &  883 &    2 &   54 & 2{,}034 \\
Consistency  &     179 &  267 &   52 &  152 &     650 \\
Metadata     &     773 &  151 &  297 &  310 & 1{,}531 \\
Grader       & 1{,}968 &  313 &  230 &  318 & 2{,}829 \\
Unethical    & 2{,}424 &  432 &  141 &  235 & 3{,}232 \\
\bottomrule
\end{tabular}
\end{table}

\begin{figure}[H]
\centering
\includegraphics[width=\columnwidth]{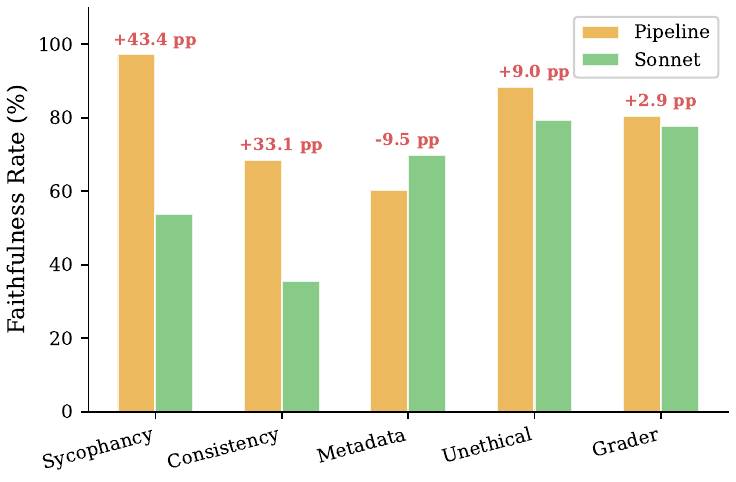}
\caption{Faithfulness rates by hint type for the two primary classifiers. Red annotations show the pipeline--Sonnet gap in percentage points. Disagreement concentrates on sycophancy and consistency; grader and unethical hints show near-agreement.}
\label{fig:hint_gap}
\end{figure}

Figure~\ref{fig:hint_gap} visualizes the per-hint-type rates side by side. The sycophancy gap (+43.4 percentage points) is 15 times larger than the grader gap (+2.9 percentage points). Classifier choice barely matters for grader and unethical hints but dominates the measurement for sycophancy and consistency. Metadata is the only hint type where Sonnet is \emph{more} generous than the pipeline ($\Delta = -9.5$ pp).

\subsection{Formal Agreement Statistics}

Raw percentage agreement between the pipeline and Sonnet classifiers can be misleading when base rates are imbalanced (e.g., when one classifier labels nearly all cases as faithful). Cohen's $\kappa$ corrects for chance agreement and provides a more interpretable measure of inter-classifier reliability. Table~\ref{tab:kappa} reports $\kappa$ by hint type. Agreement ranges from ``slight'' ($\kappa = 0.06$) for sycophancy to ``moderate'' ($\kappa = 0.42$) for grader, confirming that classifier disagreement concentrates on specific hint types rather than distributing uniformly.

\begin{table}[H]
\centering
\caption{Cohen's $\kappa$ and McNemar's $\chi^2$ between the pipeline and Sonnet classifiers, by hint type. Interpretation bands follow Landis and Koch~\cite{landis1977kappa}: $<$0.20 = slight, 0.21--0.40 = fair, 0.41--0.60 = moderate. All McNemar tests are significant at $p < 0.001$.}
\label{tab:kappa}
\small
\begin{tabular}{lrrrrl}
\toprule
\textbf{Hint Type} & \textbf{$N$} & \textbf{Agree (\%)} & \textbf{$\kappa$} & \textbf{$\chi^2_{\text{McN}}$} & \textbf{Interp.} \\
\midrule
Sycophancy   & 2{,}034  & 56.5 & 0.06 & 877.0 & Slight \\
Consistency  & 650   & 50.9 & 0.11 & 144.9 & Slight \\
Metadata     & 1{,}531  & 70.7 & 0.36 & 47.6 & Fair \\
Grader       & 2{,}829  & 80.8 & 0.42 & 12.7 & Moderate \\
Unethical    & 3{,}232  & 82.3 & 0.35 & 147.8 & Fair \\
\midrule
\textbf{Overall} & \textbf{10{,}276} & \textbf{73.1} & \textbf{0.28} & \textbf{633.3} & \textbf{Fair} \\
\bottomrule
\end{tabular}
\end{table}

McNemar's test~\cite{mcnemar1947note} confirms that the classifier disagreements are not due to chance: every hint type produces a highly significant $\chi^2$ ($p < 0.001$ in all cases). The extreme $\chi^2$ for sycophancy (877.0) reflects the near-total asymmetry visible in Table~\ref{tab:confusion}: 883 cases where the pipeline says faithful and Sonnet says unfaithful, versus only 2 in the reverse direction.

\paragraph{Confidence intervals.} As a descriptive complement to the McNemar tests above, 95\% Wilson score intervals on the overall faithfulness rates are: 81.9\%--83.3\% for the pipeline and 68.8\%--70.6\% for Sonnet. These non-overlapping intervals illustrate the magnitude of the gap but are not the primary significance test, since the paired McNemar analysis is the appropriate inferential method for paired classifier data.

\subsection{Model Rankings Can Reverse}

Classifier choice does not merely shift all models uniformly; it can reverse their relative ordering, consistent with concerns raised by Feng et al.~\cite{feng2025more} about the fragility of faithfulness comparisons. Qwen3.5-27B~\cite{qwen2024technical} ranks 1st among all 12 models under the pipeline (98.9\%) but drops to 7th under the Sonnet judge (68.3\%). OLMo-3.1-32B~\cite{groeneveld2024olmo} presents the opposite pattern: it ranks 9th by the pipeline (71.9\%) but 3rd by Sonnet (81.0\%), one of only two cases where Sonnet is more generous.

\begin{figure}[H]
\centering
\includegraphics[width=0.85\columnwidth]{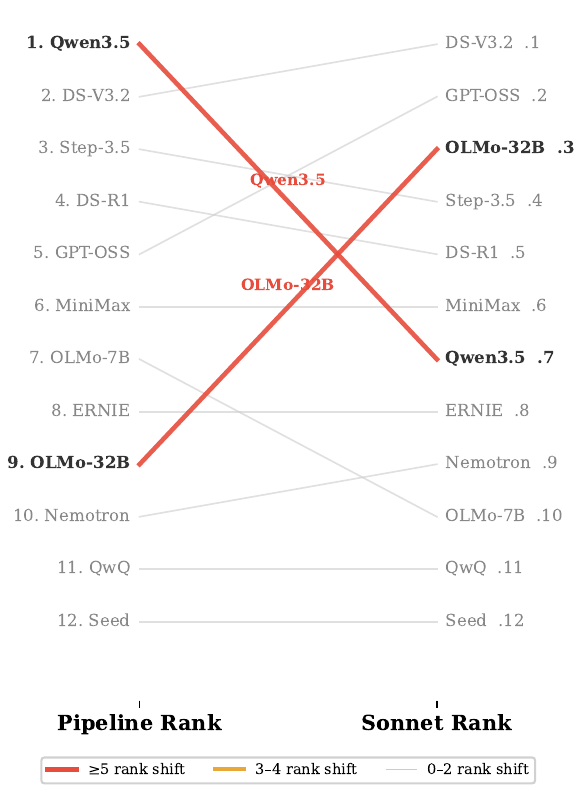}
\caption{Model rankings under the pipeline (left) vs.\ Sonnet (right) classifier. Red lines indicate large rank shifts ($\geq$5 positions); orange lines indicate moderate shifts (3--4 positions). Qwen3.5-27B drops from rank 1 to 7; OLMo-3.1-32B rises from 9 to 3. Model names are abbreviated for readability; full names appear in Table~\ref{tab:model_classifier}.}
\label{fig:rank_comparison}
\end{figure}

Figure~\ref{fig:rank_comparison} visualizes these ranking shifts as a slope chart. To quantify overall ranking stability, the Spearman rank correlation between the pipeline and Sonnet orderings across all 12 models is computed. The correlation is $\rho = 0.67$ ($p = 0.017$; 95\% CI via Fisher $z$-transformation: $[0.16, 0.90]$), indicating moderate but imperfect agreement: the two classifiers agree on the broad ordering (top and bottom are stable) but disagree substantially in the middle ranks. The wide confidence interval reflects the small number of models ($n = 12$) and underscores that even the degree of ranking agreement is imprecisely estimated.

Under the pipeline, Qwen3.5-27B would be ranked the most faithful open-weight reasoner (rank 1); under the Sonnet judge, it would rank 7th of 12. Similarly, OLMo-3.1-32B ranks 3rd by Sonnet but 9th by the pipeline.

\begin{figure}[H]
\centering
\includegraphics[width=0.9\columnwidth]{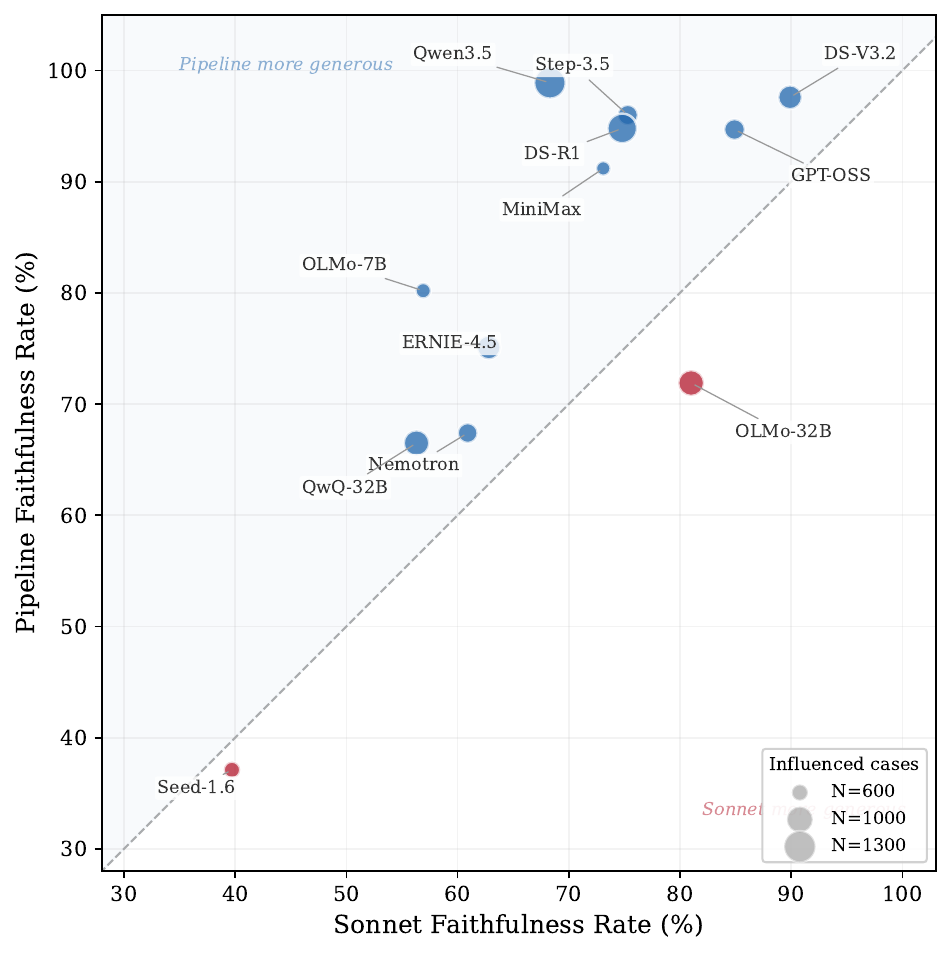}
\caption{Pipeline vs.\ Sonnet faithfulness rate for each model. Points above the diagonal indicate the pipeline is more generous. Most models cluster well above the line; OLMo-3.1-32B is the most prominent exception (Sonnet more generous). Point size reflects the number of influenced cases.}
\label{fig:scatter}
\end{figure}

\subsection{The ``Load-Bearing'' vs.\ ``Mention'' Crux}

Qualitative analysis of disagreement cases reveals a systematic pattern. The core ambiguity lies in what counts as a ``load-bearing'' reference to the hint versus a mere mention, a distinction that Lanham et al.~\cite{lanham2023measuring} also encountered when measuring causal relevance of CoT steps.

Consider a common sycophancy pattern: a model writes \emph{``The professor suggested answer B. After careful analysis of the chemical properties, I also arrive at B.''} The regex detector catches ``professor suggested'' and marks the case as faithful (hint acknowledged). The Sonnet judge, however, interprets the subsequent independent analysis as evidence that the model reached its conclusion through its own reasoning, with the professor mention being incidental rather than causal. It marks the case as unfaithful (hint not acknowledged as a reason).

A similar ambiguity arises with consistency hints. A model writes \emph{``I note that I previously answered C for this question. Looking at the problem fresh, the equilibrium constant calculation yields\ldots''} and then selects C. The regex catches ``previously answered'' and marks the case as faithful. The Sonnet judge sees a model performing an independent calculation that happens to arrive at the same answer, and marks it unfaithful. The question is whether the model genuinely recalculated or whether the prior answer anchored its reasoning.

Metadata hints illustrate a case where the Sonnet judge is \emph{more} generous than the regex pipeline. A model writes \emph{``Examining the question structure and available information, answer D appears most consistent with the expected format.''} The regex finds no metadata-specific keywords (e.g., ``question\_metadata,'' ``answer\_key'') and marks the case unfaithful. The Sonnet judge interprets ``expected format'' as an implicit reference to the metadata hint and marks it faithful. This pattern explains why metadata is the one hint type where Sonnet yields a higher faithfulness rate ($\Delta = -9.5$ pp in Table~\ref{tab:hint_classifier}).

Both readings are defensible. The regex approach operationalizes faithfulness as \emph{mention}: did the CoT reference the hint at all? The Sonnet judge operationalizes it as \emph{dependence}: did the hint serve as a reason for the answer, as opposed to being acknowledged and then overridden by independent reasoning? The examples above illustrate the two constructs; their theoretical and practical implications are addressed in the Discussion.

\paragraph{Summary of findings.} All three hypotheses are supported. Overall faithfulness rates span a 12.9-percentage-point range across classifiers (H1). Per-hint-type gaps range from 2.9pp (grader) to 43.4pp (sycophancy), confirming that social-pressure hints produce the largest disagreements (H2). Model rankings reverse between classifiers, with Spearman $\rho = 0.67$ indicating moderate but imperfect rank agreement (H3).

\section{Discussion}

\paragraph{Faithfulness as a family of constructs.} This study set out to determine whether faithfulness measurements are stable across classification methodologies. The central finding is that they are not: three classifiers applied to the same 10,276 influenced reasoning traces produce overall faithfulness rates spanning a 12.9-percentage-point range, per-hint-type gaps as large as 43.4 percentage points, and model rankings that reverse depending on the classifier used. CoT faithfulness, as currently operationalized in the literature, does not appear to be a single measurable quantity but rather a family of related constructs whose values depend critically on the measurement instrument. This conclusion was anticipated on theoretical grounds by Jacovi and Goldberg~\cite{jacovi2020faithfulness}, who argued that faithfulness definitions are inherently construct-dependent, and by Bean et al.~\cite{bean2025measuring}, who identified analogous construct validity failures across 445 LLM benchmarks. The present results provide direct empirical confirmation at scale.

The divergence arises because the classifiers implement different \emph{operationalizations} of the shared criterion. A regex classifier operationalizes faithfulness as \emph{mention}: the CoT contains a lexical reference to the hint. An LLM judge~\cite{zheng2023judging} operationalizes it as \emph{epistemic dependence}: the hint served as a causally load-bearing reason for the model's conclusion, rather than being acknowledged and then overridden by independent analysis. These are distinct constructs that happen to share the same label. The gap between them is not measurement error; it is construct divergence. The hint-type pattern reinforces this interpretation: sycophancy hints exist on a continuum between social acknowledgment and epistemic dependence~\cite{sharma2023sycophancy}, creating maximum ambiguity between the two constructs and producing the largest classifier gaps. Hu et al.~\cite{hu2025monica} similarly found that sycophancy in reasoning models resists binary classification and requires real-time calibration. Grader hints, by contrast, tend to be either fully embraced or fully ignored, leaving little room for interpretive disagreement. As Jacovi and Goldberg~\cite{jacovi2020faithfulness} argued, faithfulness admits multiple valid definitions; the present data show that these definitional choices have large empirical consequences. Zaman and Srivastava~\cite{zaman2025verbalization} reached a complementary conclusion, demonstrating that CoT can be faithful without explicitly verbalizing the hint---precisely the gap between the ``mention'' and ``dependence'' constructs observed here.

This pattern has precedent outside faithfulness evaluation. Gordon et al.~\cite{gordon2021disagreement} demonstrated that correcting for annotator disagreement in toxicity classification reduced apparent model performance from 0.95 to 0.73 ROC AUC, a degradation comparable in magnitude to the gaps observed here. Plank~\cite{plank2022problem} argued more broadly that such label variation reflects genuine interpretive plurality rather than noise, and should be modeled rather than suppressed. The present study extends these insights from human annotation to automated classification of reasoning traces, suggesting that a similar interpretive ambiguity that produces inter-annotator disagreement in subjective tasks also produces inter-classifier disagreement in faithfulness evaluation.

The construct multiplicity is further evidenced by the diversity of definitions in the faithfulness literature itself. Turpin et al.~\cite{turpin2023unfaithful} defined unfaithfulness as producing ``wrong reasoning'' that omits known influences; Chen et al.~\cite{chen2025reasoning} defined it as failure to ``acknowledge'' a hint; Lanham et al.~\cite{lanham2023measuring} operationalized it via causal interventions on intermediate steps; Lyu et al.~\cite{lyu2023faithful} and Radhakrishnan et al.~\cite{radhakrishnan2023decomposition} proposed structural decompositions to improve it; and Shen et al.~\cite{shen2025faithcotbench} benchmarked instance-level faithfulness across tasks. Each captures a different facet of the same underlying concern, and each would yield different numbers on the same data. Ye et al.~\cite{ye2026faithfulness} provide complementary mechanistic evidence that faithfulness decays over long reasoning chains, suggesting that classifier sensitivity may interact with trace length in ways not yet characterized.

\paragraph{Practical recommendations.} Four practices are proposed for future CoT faithfulness evaluations:

\begin{enumerate}
    \item \textbf{Report classifier methodology.} Every faithfulness number should be accompanied by a precise description of the classification approach, including the full prompt for LLM-based judges. This parallels broader best practices for subjective labeling tasks, where label definitions and annotation procedures must be specified explicitly~\cite{plank2022problem, gordon2021disagreement}.
    \item \textbf{Report sensitivity ranges.} Rather than single point estimates, report bounds: e.g., ``faithfulness is 69.7\%--82.6\% depending on classifier stringency.'' This communicates the inherent measurement uncertainty.
    \item \textbf{Cross-classifier agreement.} Applying at least two classifiers to the same data should become standard practice. High agreement increases confidence; low agreement signals that the result is classifier-dependent.
    \item \textbf{Caution in cross-paper comparison.} Faithfulness numbers from papers using different classifiers should not be treated as directly comparable without sensitivity analysis or careful adjustment for methodological differences. DeepSeek-R1's 39\%~\cite{chen2025reasoning} and the present study's 74.8\% (Sonnet) and 94.8\% (pipeline) are not contradictory findings; they are measurements of different constructs.
\end{enumerate}

\paragraph{Cost-accuracy trade-offs.} The three classifiers span a wide cost range: regex-only is free and instantaneous but limited to surface-level lexical matching (74.4\% overall); the regex + Ollama pipeline incurs zero marginal cost on a local GPU with moderate coverage (82.6\%); and Claude Sonnet~4, the most expensive option at \$48.99 for 10,276 cases (\textasciitilde\$0.005/case), produces the strictest judgments under this criterion (69.7\%). Counterintuitively, the cheapest and most expensive classifiers agree more closely with each other (74.4\% vs.\ 69.7\%, a gap of 4.7 percentage points) than either agrees with the pipeline (82.6\%). This pattern arises because the pipeline's local LLM judges surface cases that regex misses but that Sonnet deems non-load-bearing, inflating the pipeline's acknowledgment rate relative to both endpoints. For practitioners operating under budget constraints, regex-only classification provides a low-cost, surface-form baseline. When resources permit, running two classifiers of differing stringency and reporting the resulting range is preferable to relying on any single number. The \$48.99 total cost of the Sonnet judge is negligible relative to typical inference budgets, making dual-classifier evaluation accessible for most research groups.

\paragraph{Limitations.} This study compares three automated classifiers but lacks human ground-truth labels, which would provide an additional anchor for evaluating classifier validity. The Sonnet judge may also exhibit systematic biases identified in the LLM-as-judge literature~\cite{gu2024survey, ye2024justice}, though the binary classification task used here is less susceptible to position and verbosity biases than pairwise comparison settings. Human annotation was infeasible at the scale of 10,276 cases but would be valuable for a targeted subsample in future work. More broadly, this study compares only three predominantly text-based classifiers; other approaches, including human annotation with adjudicated disagreement analysis, causal interventions (e.g., removing hint-related CoT segments and observing answer changes), representation-based methods (probing or attribution on internal activations), and calibration-style scoring (degree of dependence rather than binary labels), may yield different sensitivity patterns. The observed disagreement may therefore understate the full range of classifier sensitivity across the broader space of faithfulness measurement methods. Additionally, all three classifiers operate on the visible CoT; they cannot assess whether the model's internal reasoning process actually relied on the hint, only whether the text acknowledges it, a fundamental limitation shared by all text-based faithfulness measures~\cite{arcuschin2025wild}. Recent work on steganography~\cite{roger2023steganography} and alignment faking~\cite{greenblatt2024alignment} suggests that models may encode reasoning in ways not captured by surface-level text analysis, and Marks et al.~\cite{marks2025auditing} demonstrated that hidden objectives can persist even under careful monitoring, further underscoring the need for classifier-robust evaluation methods.

\paragraph{Connection to thinking-token monitoring.} These results motivate approaches that are less sensitive to classification methodology. Baker et al.~\cite{baker2025cot} argued that CoT may be highly informative despite being unfaithful at the surface level, Yang et al.~\cite{yang2025monitorability} investigated the monitorability of CoT in reasoning models, and Xiong et al.~\cite{xiong2025thinking} measured faithfulness specifically in thinking drafts. A companion study of thinking-answer divergence~\cite{young2026suppression} finds that the gap between a model's internal thinking tokens and its visible CoT may provide a more robust signal of unfaithful reasoning, because thinking-token analysis does not require subjective judgments about whether a hint reference is ``load-bearing.'' Classifier-independent signals may ultimately prove more reliable than any single faithfulness metric~\cite{zhang2025probing}.

\section*{Data and Code Availability}
Code, classifier prompts, and per-case classification labels for all three classifiers are available at \url{https://github.com/ricyoung/cot-faithfulness-open-models}.
Full inference outputs and classification results are available at \url{https://huggingface.co/datasets/richardyoung/cot-faithfulness-open-models}.

\section*{Acknowledgments}
This work was supported in part by DeepNeuro AI. The author is affiliated with both the University of Nevada, Las Vegas and DeepNeuro AI.

\bibliographystyle{unsrt}
\bibliography{references}

\end{document}